\documentclass[runningheads]{llncs}
\usepackage{graphicx}
\usepackage{xcolor}

\usepackage{geometry}
\usepackage{hyperref}
\hypersetup{
    linkcolor=blue,
    filecolor=blue,      
    urlcolor=blue,
    citecolor=blue,
    colorlinks=true,
    pdftitle={Three Conjectures on Unexpectedess},
    pdfpagemode=FullScreen
}

\usepackage{amsmath}
\usepackage{amsfonts}
\usepackage{amsthm}
\newtheorem{conj}{Conjecture}
\usepackage[bb=dsserif]{mathalpha}

\begin{document}
\title{Three Conjectures on Unexpectedeness}
%
%
\author{Giovanni Sileno\inst{1} \and
Jean-Louis Dessalles\inst{2}}
\authorrunning{G.~Sileno and J.-L.~Dessalles}
%
\institute{University of Amsterdam, Amsterdam, the Netherlands\\ \and
Télécom Paris, Institut Polytechnique de Paris, Paris, France\\
\email{g.sileno@uva.nl}, \email{dessalles@telecom-paris.fr}}
\maketitle              
\begin{abstract}
Unexpectedness is a central concept in Simplicity Theory, a theory of cognition relating various inferential processes to the computation of Kolmogorov complexities, rather than probabilities. Its predictive power has been confirmed by several experiments with human subjects, yet its theoretical basis remains largely unexplored: \textit{why} does it work? This paper lays the groundwork for three theoretical conjectures. First, unexpectedness can be seen as a generalization of Bayes' rule. Second, the frequentist core of unexpectedness can be connected to the function of tracking ergodic properties of the world. Third, unexpectedness can be seen as constituent of various measures of divergence between the entropy of the world (environment) and the variety of the observer (system). The resulting framework hints to research directions that go beyond the division between probabilistic and logical approaches, potentially bringing new insights into the extraction of causal relations, and into the role of descriptive mechanisms in learning.

\end{abstract}
\section{Introduction}

Shannon's theory of information \cite{Shannon1948} entails that a uniform source of noise is maximally informative,  
but very few humans would agree with such a conclusion. 
Motivated by addressing this shortcoming, \textit{simplicity theory} (ST) has been presented as a theory of cognition explaining various phenomena observed in experiments with humans concerning \textit{relevance} \cite{Dessalles2008,Dessalles2011,Dessalles2011a,Dessalles2013}. Theoretically, ST builds upon \textit{algorithmic information theory} (AIT), and focuses on the measure of \textit{unexpectedness} ($U$), resulting from the difference of two (bounded) Kolmogorov complexities, computed on two distinct machines. Examples of phenomena correctly predicted by ST include: remarkable lottery draws (e.g. $11111$ is more unexpected than $64178$, even if the lottery is fair), coincidence effects (e.g. meeting by chance a friend in a foreign city is more unexpected than meeting any unknown person equally improbable), deterministic yet unexpected events (e.g. a lunar eclipse), and many others 
\cite{Dessalles2008,Dessalles2011,Dessalles2011a,Dessalles2013}. These experiments show that unexpectedness has predictive power, yet, its theoretical underpinnings remain largely unexplored: \textit{why does unexpectedness function}?

We will therefore present in this document three ``conjectures'' on unexpectedness. The first conjecture (section 3) highlights a strong correspondence between ST's formulation of unexpectedness and Bayes' rule; we summarize here what has been presented more extensively in \cite{Sileno2022}. This formal mapping can be useful for practical purposes: all applications which are based on Bayes' rule could be generalized using unexpectedness. Yet, the mapping does not offer further insight into the origins of the unexpectedness, as a sound cognitive function would. Therefore, we formulate here two additional hypotheses. In section 4, we observe that, focusing on its ``frequentist'' core, unexpectedness can be seen as a \textit{measure of change of ergodic properties of the world}.  In section 5 we recover the descriptive dimension, and elaborate on its crucial role in the \textit{interactions between environmental entropy and system variety}. For completeness, section 2 provides a concise overview of the fundamental concepts relevant to our elaboration, aiming to make this document as much as complete as possible; knowledgeable readers are invited to pass through it, and to refer to it only at need.

\section{Fundamental concepts}

\subsection{Probability theory}

Probability theory was historically introduced to study games of chance, and thus dealing with events that occur in a finite set called \textit{sample space}, usually denoted with $\Omega$. 
This core branch, named \textit{discrete probability theory}, assumes that for each element $\omega \in \Omega$ we have an atomic ``probability'' value $p(\omega)$, which satisfies the following properties:
$$\forall \omega \in \Omega: p(\omega)\in [0,1] \qquad \sum _{\omega\in \Omega }p(\omega)=1$$
A (random) \textit{event} is then defined as any subset $E$ of the sample space $\Omega$. This definition allows us to specify the set $E$ using logical expressions that correspond to set operations in set theory.
The \textbf{probability} of the event $E$ is then defined as:
$$P(E)=\sum _{\omega \in E}p(\omega)$$
The probability of the event coinciding with the entire sample space is $1$, and the probability of the null event is $0$. The function $p(\omega)$ mapping a single point in the sample space to its atomic ``probability'' value is called \textit{probability mass function}. Note that this formulation does not tell us how to obtain a probability mass function, but assumes its existence. 

Given this basis, a \textit{random variable} $X$ is defined as a function (eg. $x = f(E)$, with $E \subseteq \Omega$) that assigns (typically real) values to each possible outcomes (probabilistic event) generated through an extraction from the sample space. Introducing the notion of random variables is necessary to compute  \textit{expected values} as the average of all possible outcomes (suppose collected in a set $\xi$), weighted by their probability:
$$\mathbb{E}[X]=\sum_{E \in \xi} P(E) \cdot f(E)$$

Focusing on repeated extractions, a \textit{random process} $\{X_n\}$ is defined as a sequence of random variables ($X_1, \ldots,  X_n$), where the index of the sequence expresses a temporal coordinate.

\subsection{Information theory}

The contribution that Shannon's theory of information \cite{Shannon1948} has had in enabling contemporary digital technologies cannot be underestimated. Yet, it is often overlooked that the perspective through which it was introduced was a specific one: the \textit{transmission} of information. 
Taking Shannon's standpoint, the information of a  certain object relates only to the (communication) source from which this object is extracted (transmitted), regardless of the actual ``content'' of the object. This theoretical approach is what enables us to define the \textit{information} of a symbol $x$ which appears with probability $P(x)$ as:
$$I(x) = \log \frac{1}{P(x)}$$

The \textbf{entropy} of a certain source is then given by the average information across all symbols of that source, weighted by their probability of occurrence, ie. the expected value of the information of a random variable $X$ associated to the source, defined over the set of symbols $\Omega$ with probability $P$:

$$H(P) = \sum_{x \in \Omega} P(x) I(x) = \sum_{x \in \Omega} P(x) \log \frac{1}{P(x)} = \mathbb{E}[I(X)]$$

Given two discrete probability distributions $P$ and $Q$ defined on the same sample space $\Omega$, the ``relative entropy'' of $P$ (the ``actual'' distribution) from $Q$ (the ``model'' distribution) can be defined via the \textit{Kullback-Leibler divergence}:
\begin{align*}
\mathbb{D}_{\text{KL}}(P \parallel Q) &= \sum_{x\in{\Omega}} P(x) \log \frac{P(x)}{Q(x)} \\ &= \sum_{x\in{\Omega}} P(x) \left[ \log \frac{1}{Q(x)} - \log \frac{1}{P(x)} \right] = H(P, Q) - H(P)
\end{align*}
where $H(P, Q)$ is the \textbf{cross-entropy}:
$$H(P, Q) = \sum_{x\in{\Omega}} P(x) \log \frac{1}{Q(x)}$$

\subsection{Variety}

The concept of \textbf{variety} has been introduced in cybernetics by Ashby \cite{Ashby1956} in relation to the set of distinguishable states $S$ expressed by the system under focus, and is measured either in linear or logarithmic forms. We will consider here the second formulation:
$$ V = \log_2|S| $$
One well-known result of cybernetics is the \textit{law of requisite variety}, a necessary but not sufficient criterion for control, which informally can be expressed as:
\begin{quote}\textit{The variety of the system needs to match the variety of the environment for the system to be able to control the environment.}\end{quote} 
Note that the the concept of variety does not involve or refer to underlying probabilities. However, since its introduction, a strong connection has been found with Shannon's information theory.

\subsection{Algorithmic Information Theory}

The modern formalization of probability theory has been set by Kolmogorov. Kolmogorov, together with Solomonoff and Chaitin, is also one of the founding fathers of \textit{algorithmic information theory} (AIT) (see eg. the overview given in \cite{Li2008}), a theoretical framework that provides an approach to information orthogonal to Shannon: what an object conveys is expressed by the object itself, rather than by the source from which it is generated (extracted). In parallel and independent efforts, 
focusing on simple computational objects as strings, these three authors converged to a principle that informally can be described as:
\begin{quote}
\textit{Strings are simpler if they have shorter descriptions.
}\end{quote}
Eventually named after Kolmogorov, the \textbf{Kolmogorov complexity} of a string $x$ is defined as the minimal length of a program that, given a certain optional input parameter $y$, produces $x$ as an output: 
$$K_\phi(x|y) = \min_p \big\{|p| : p(y) = x \big\}$$ 
The length of the minimal program depends on the operators and symbols available to the machine $\phi$ running the program.\footnote{Note that $K$ is an \textit{algorithmic informational complexity}: it captures how much information is needed for constructing the object, but not how much time or space is required. For this reason, it is distinct from the most commonly used \textit{algorithmic/time complexity}, relevant to study tractability.} If specified on universal Turing machines, this measure is generally \textit{incomputable} (essentially due to the halting problem), and it is defined always up to a constant. If the machine is resource-bounded, however, complexity is computable. The bounded version will be denoted as $C$. 

Note that this definition of complexity can be mapped to  domains other than strings, as long as one defines what the \textit{objects} are and the \textit{operations} that can be performed on these objects. Furthermore, under certain conditions, the search for the minimal program can be realized via \textit{min-path} or functionally similar algorithms.

\subsection{Unexpectedness in Simplicity Theory\label{sec:unexpectedness-in-ST}}

Simplicity Theory (ST) stems from the empirical observation that, for humans, 
\begin{quote}
\textit{Situations are unexpected if they are simpler to describe than to explain.}
\end{quote}
where explanations here map to causal chainings producing the situation in focus. Mathematically, and building upon AIT, \textbf{unexpectedness} is measured as a \textit{drop of complexity}:
$$ U(s) = C_W(s) - C_D(s) $$
where $C_W$ and $C_D$ are Kolmogorov complexities, computed via distinct machines. A diagrammatic representation of the domains of the two complexities can be expressed as:
$$\overbrace{\textrm{world} \rightarrow \textrm{situation} }^{C_W} \qquad \overbrace{\textrm{situation} \leftarrow \textrm{mind}}^{C_D}$$
The \textit{causal complexity} $C_W$ (also \textit{world complexity} or \textit{generation complexity}) builds upon a world model maintained by a world machine $W$ (whose operators typically  concern occurrences, causal dependencies, causal compositionality, \dots). The \textit{description complexity} $C_D$ builds upon a mind model maintained by a description machine $D$ (whose operators concern concept retrieval, association, and various forms of compositionality and concept transformation, eg. repetition, contrast, etc.). Because complexities are expressed on a logarithmic  scale, one may introduce an additional constraint:
$$ U(s) \ge 0 $$ 
to capture a principle of \textit{cognitive economy}: situations are described up to the extent they are unexpected to occur.     

The following sections will present three different conjectures on the theoretical underpinning justifying unexpectedness as a sound measure, and attempting to explain its emergence as a cognitive mechanism.

\section{Unexpectedness and Bayes' rule}

\begin{conj}Bayes’ rule is a specific instantiation of a more general template captured in ST by Unexpectedness.
\end{conj}

\noindent To construct this claim, we start from the definition of \textit{conditional probability}: 
$$P(O \cap M) = P(M|O) \cdot P(O) = P(M) \cdot P(O|M)$$
where $O$ denotes an observation, and $M$ a model (both elements from the same measurable space). Bayes' formula is easily derived:
$$P(M|O) = \frac{P(M \cap O)}{P(O)} = \frac{P(O|M) \cdot P(M)}{P(O)}$$
The formula is often expressed using informal terms: 
$$\textrm{posterior} = \frac{\textrm{likelihood} \cdot \textrm{prior}}{\textrm{evidence}}$$
Empirical observations \cite{Dessalles2011} suggest that unexpectedness $U$ can be put in correspondence to posterior probability, i.e.
$$\textrm{posterior} = 2^{-U}$$ 
This entails that when $U \approx 0$ ($\textrm{posterior} \approx 1$), the situation confirms the agent's model of the world (it is ``plausible''), and therefore it is not informative.\footnote{Note that, to maintain a correspondence with probabilities, $U$ needs also to be superior or at least equal to 0. This constraint maps to the principle of cognitive economy expressed in section \ref{sec:unexpectedness-in-ST}.} 

However, in applying this mapping, we tacitly overlooked a detail. Unexpectedness has only a parameter $s$, whereas posterior probability refers to $O$ and $M$. Intuitively, $s$ (the situation in focus) corresponds to $O$ (the observation) and not to $M$. But then, where can we find $M$? In order to understand this absence, let us reconsider Bayes' formula. Inverting the terms of the equation, and using the logarithm, we can form a mapping to unexpectedness, i.e.:
$$\overbrace{\log \frac{1}{P(M|O)}}^{U(s)} = \log \frac{P(O)}{P(O|M)\cdot P(M)} = \overbrace{\log \frac{1}{P(O|M)} + \log\frac{1}{P(M)}}^{C_W(s)} - \overbrace{\log\frac{1}{P(O)}}^{C_D(s)}$$
We can then analyze the suggested correspondences individually. 

\subsection{Role of causal complexity}

Let us start from $C_W(s)$, the causal complexity, i.e. the length in bits of the shortest path that, according to the agent's world model, generates the situation $s$. If the situation $s$ concerns a phenomenon --- an \textit{event} probabilistically captured by $O$ --- $s$ can be seen as the manifestation of some pre-existing generative context $c$, that probabilistically is captured with $M$. Then, in order to generate $s$ (e.g. the symptoms of a disease), the world has first to generate its cause $c$ (e.g. the disease), a passage expressing the application of a \textit{chain rule}:
$$C_W(s) \leadsto C_W(c * s) = C_W(s||c) + C_W(c)$$ 
where $C_W(s||c)$ is the complexity of generating $s$ from a state of the world in which $c$ is the case, and $c * s$ is the sequential chaining of $c$ and $s$ ('$||$' and '$*$' add temporal constraints that '$|$' and '$\cap$' in probability formulas do not have). From the definition of Kolmogorov complexity, the mapping is an equality if and only if the shortest path to $s$ passes from $c$, i.e. if $c$ is the \textit{best explanation} of $s$:
$$C_W(s) = \min_{c} C_W(c * s) = \min_{c} \left[ C_W(s||c) + C_W(c) \right ]$$
Therefore the unexpectedness formula can be seen as abstracting the \textit{causally explanatory} factor $c$, with the implicit assumption that the best one is automatically selected in the computation of complexity. 

\subsection{Role of description complexity}

ST specifies $C_D$, the description complexity, as the length in bits of the shortest determination of the object $s$. Such shortest determination may consist e.g. in specifying the address where to retrieve it from memory.\footnote{Note that from a computational point of view, $U$ could be negative, namely when the description of $s$ is more complex than its generation; we are in this case in front of \textit{inappropriate} descriptions, as they are adding irrelevant information for their function.} In the terms suggested by Bayes' formula, $C_D$ corresponds to the probability of having \textit{observed} a certain situation. The link between descriptive complexity and probability can be then established through \textit{optimal encoding} following information theory (the length of the encoding associated to $O$ should approximate $\log\frac{1}{P(O)}$), where probability is assessed through frequency. However, this approach does not take into account possible mental compositional effects (e.g. repetition, which can be related to Gestalt-like phenomena), nor events that never occurred before. Complexity is a more generally applicable measure than probability.

\subsection{Comparison with Bayes' rule}

The previous observations allow us to claim that Bayes' rule is a specific instantiation of ST's Unexpectedness that: (a) makes a candidate ``cause'' explicit and does not select automatically the best candidate; (b) takes a frequentist-like approach for encoding observables. More formally: 
$$U(s) = \min_c \overbrace{ \left[ C_W(c * s)  - C_D(s) \right]}^{\textrm{posterior}} = \min_c [ \overbrace{C_W(s||c)}^{\textrm{likelihood}} + \overbrace{C_W(c)}^{\textrm{prior}}  - \overbrace{C_D(s)}^{\textrm{evidence}}]$$
Note that this formula relies on the explicit assumption that $c$ precedes $s$ (as indicated by the symbols $*$ and $||$). This restriction is absent from Bayes' rule, in which the model $M$ and the observation $O$ can exchange roles; their causal dependence does not lie in the rule, but solely in the eye of the modelers.

\section{Unexpectedness and Ergodicity}

\begin{conj}Unexpectedness, in its frequentist core, tracks changes in the ergodicity of the environment.
\end{conj}

Intuitively, how concepts are created and maintained in the agent's mind is not independent of how the world manifests itself to the agent. Additionally, because the agent does not have access to the actual physical laws, $C_W$ has to rely---at least in part---upon $C_D$. What is then the relation between causal and description complexity? And what is the relation of the two complexities with experience? 

Simplicity Theory considers $C_W$ and $C_D$ to be associated to two bounded Turing machines with distinct operators. $C_W$ may employ causal chaining. $C_D$ may employ various descriptive operators (eg. repetition). Before referring to compositional operations, however, complexities derive also from primitive resources, such  as the strengths of causal links $C_W(s||c)$, of conceptual accessibility $C_D(s|c)$, and the complexity of individual instances, concerning occurrences $C_W(s)$, and retrieval $C_D(s)$. In other words, although chaining may be used to compute complexities when no previous information of a certain situation is available, we still need some way to ground perceptual experience.
 
Focusing on non-compositional primitives, and in particular on those concerning atomic situations, the problem can be reduced to investigating the interplay between \textit{frequency} (for the generative aspect) and \textit{recency} (for the accessibility aspect). We will now show that, for this frequentistic core, the world complexity $C_W$ functions as an estimator of ergodic processes of growth, and that the unexpectedness $U$ can be interpreted as a signal associated to a meta-cognition function, measuring to what extent that estimator is (not) functioning. 

\subsection{Background: Ergodicity and Stationarity}

A random process is said \textbf{ergodic}, if, for time sufficiently great, its statistical average is the same as its temporal average. 

Let us suppose, that during its activity, a system might be in any state $\omega \in \Omega$; the state at a certain point in time $n$ is denoted as $\omega_n$; $f(\omega_n)$ is a random variable associated to that state. Ergodicity can be captured by the Birkhoff equation (here expressed with respect to discrete time, and to a discrete random variable):
$$ \lim_{N\rightarrow\infty}\frac{1}{N}\sum_{n=1}^N f(\omega_n) = \sum_{\omega \in \Omega} f(\omega)P(\omega)$$
In other words, by taking an adequate number of samples in any point in time we can still extract some information of the underlying distribution. 

Instead, a random process is said \textit{stationary} if the probability distribution of the associated random variable does not change in time. Stationarity can also be defined up to a certain probabilistic moment (e.g. mean for the first, variance for the second, etc.).

The definition of ergodicity implicitly assumes stationarity of the process in the first moment, because the time average needs to converge to the mean of the distribution, a requirement plausible only if the mean is constant. Not all stationary process are ergodic though. This scenario occurs especially when there are \textit{bifurcation points}, such as those determined by superposition of independent processes with different temporal characteristics.\footnote{For instance, consider the random process $\{Y_n\}$ defined as $Y_n=X_n+V$, where $\{X_n\}$ is stationary and ergodic, and $V$ is a stationary random variable extracted only once. $\{Y_n\}$ is stationary (the statistics remains unvaried in time), but it is non-ergodic (the average across distributions is not the same as the average across time, as the latter depends on the specific value extracted for $V$).} This means that, when ergodicity is not satisfied, either the underlying process has changed its first-moment statistics, or we are in front of a more complex phenomenon and we have captured a bifurcation point. 

\subsection{Ergodicity of a Reference Situation}

Instead of considering all possible events (situations), we focus now on single reference $R$. At a certain point in time, the reference might be manifest (or not). This statement results from contrasting the currently observed situation $O_n$ (observed at time $n$) with the reference situation $R$, which reduces Birkhoff's equation to:

$$ \lim_{N\rightarrow\infty}\frac{1}{N}\sum_{n=1}^N \mathbb{1} _{O_n \sim R} = P(R)$$
The left-hand of the equation is temporal, bound to observations, and works by means of a \textit{determination} test, that is, on whether $O_n \sim R$.  The right-hand is instead atemporal, and provides an information of the environment deemed to be a random source providing instances which may match with the reference $R$.

This formulation captures ergodicity in terms of normalized counting, giving a measure of \textit{asymptotic growth}, i.e. a rough, average information of the generative properties of the world with respect to the manifestation of situations going under $R$. If ergodicity is satisfied, the average over a sufficiently long period of time becomes constant, and  this constant corresponds to the probability of occurrence of that situation. 

\subsection{Ranking as Expected Position}

Let us suppose for simplicity that observations are maintained in a memory functionally working as a stack: new items are always put on top and are therefore be more easily accessed. Suppose that the probability of an event $x$ to be observed, and then to be put in top of the stack is $P(x)$. We can compute the expected position of $x$ as:
\begin{align*}
\mathbb{E}[\mathsf{pos}(x)] &= 0 \cdot P(x) + 1 \cdot P(x) (1-P(x)) + ... + n \cdot P(x) (1-P(x))^n =\\ 
&= P(x) \cdot \sum_n^{+\infty} n (1- P(x))^n
\end{align*}
Applying a known mathematical series\footnote{$\sum_{n=1}^{+\infty} n y^n = \frac{y}{(1 - y)^2}$.}, 
this can be rewritten as:
$$\mathbb{E}[\mathsf{pos}(x)] = P(x) \cdot \frac{1 - P(x)}{P(x)^2} = \frac{1 - P(x)}{P(x)} = \frac{1}{P(x)} - 1 $$
which, expressed on a logarithmic scale, becomes:
$$ \log \mathbb{E}[\mathsf{pos}(x)] = \log \frac{1 - P(x)}{P(x)} \approx \log \frac{1}{P(x)}$$
The approximation is valid the more $x$ occurs rarely, i.e. $P(x)$ approaches $0$.

\subsection{From Description to Causal Complexity}

The formulas of ergodicity and expected position written above offer an alternative reading of the relation between $C_D$ and $C_W$. 
Focusing only on the retrieval from a stack acting as a short-term memory (STM), the description complexity $C_D$ of a certain situation $x$ depends logarithmically on its position in the stack:
$$C_D^{STM}(x) = \log \mathsf{pos}(x)$$
This value can be reified in a temporal function, which is discontinuous, piecewise, and monotonically increasing in each piece: each time $x$ appears, its complexity passes abruptly to 0, otherwise it keeps increasing. Note that the position on the stack can be  processed by a cognitive system in a much easier way than maintaining the count of all observations. 

Interestingly, using the \textit{expected} position as the index for a second stack working as a long term memory (LTM) is aligned with \textit{optimal encoding} principles (\text{à la} Shannon–Fano, Huffman, etc.):
$$C_D^{LTM}(x) = \log \mathbb{E} [\mathsf{pos}(x)] \approx \log \frac{1}{P(x)}$$
However, this step presupposes that $P(x)$ is a static function, which is generally not the case, and in any case we cannot assume to have direct access to all the time series to compute the expected ranking. 
We can instead compute a temporal average across a certain window. If this window is adequately large, that would approximate the limit in the periodicity equation. Formally, this could be specified as:  
$$w_N^{(t)}(x) = \frac{1}{N}\sum_{n=t-N+1}^t \mathbb{1} _{O_n \sim x}$$
Interestingly, this function can be interpreted as a \textit{low pass} filter on $C_D^{STM}(x)$. Averaging the position is only one (a FIR, or \textit{finite impulse response} filter, specifically) of the possible implementations of low-pass filter to obtain the same outcome. For instance, we could use instead a low-pass IIR (\textit{infinite impulse response}) filter:
$$w^{(t)}_\alpha(x) = (1-\alpha) \cdot \mathbb{1} _{O_t \sim x} + \alpha \cdot w^{(t-1)}_\alpha(x)$$
This formulation requires only maintaining the values in the previous time step.\footnote{Notably, this is the same learning function used in Q-learning.} 

Independently of how it is implemented, if this function is sufficiently stable in time ($\Delta w \approx 0$), from the Birkhoff equation we know that it is capturing an ergodic phenomenon of growth, and for this reason it can be interpreted as furnishing a measure of its probability of occurrence, and thus of the expected position to be used for indexing the situation $x$ in the LTM, following optimal encoding principles. Moreover, because this information is about a generative (causal) phenomenon, it may be reused to set the causal machine:
$$C_W(x) = C_D^{LTM}(x) = \log w(x) $$


\subsection{Tracking Changes of Ergodicity}

Accepting the above, in this minimal ``frequentist'' core, unexpectedness can be rewritten as:
$$U(x) = C_W(x) - C_D(x) = C^{LTM}_D(x) - C^{STM}_D(x)$$
Unexpectedness can therefore be seen as measuring a misalignment between what is being perceptually processed and what has been mentally modeled in terms of asymptotic growth.
Under this view, it becomes a signal  \textit{tracking changes} in causal mechanisms (generators), captured through their ergodic properties. 
If unexpectedness remains positive, the assumption of ergodicity is not satisfied anymore. This entails non-stationarity or the occurrence of a bifurcation, hinting that some causal event has occurred. To re-establish its correct functioning, the system has to find a better cause: either selecting another known generative mechanism (typically via abduction), or introducing a new one (learning). In short, unexpectedness can be seen as providing a \textit{meta-cognition} function, telling us how well the estimator $C_W$ is working. 

\section{Unexpectedness, Entropy, and Variety}

\begin{conj}Unexpectedness, at systemic level, captures a divergence between the (generative) entropy of the world and the (descriptive) variety of the observer.\end{conj}

Let us suppose that all environments generate discrete events $i$ from a sample set $S$ with probability $p_i$. The environment with maximal entropy will be one in which signals appear following a uniform distribution, or,  equivalently, one in which entropy coincides with variety:
$$H_{\max} = \sum_{i \in S} \frac{1}{|S|} \log \frac{1}{p_i} = \log|S| = V$$
So far, the only mutual relationship between objects is to belong to the same set $S$. In some cases, however, objects come with a given ordering, eg. when they are numbers, or letters in the alphabet. The ordering allows computing a ``distance'' between objects, as for instance the number of steps necessary to pass from one symbol into another one. Strangely enough, the traditional definition of variety does not take that into account this possibility, as it is transparent to how states are encoded. But why? Before proceeding into the conjecture, we will consider the definition of variety with respect to two basic data structures.

\subsection{Complexity of retrieval}

The cost of computing a single computational object, in the absence of operators other than retrieval, can be seen as the cost of specifying its memory address. 

\subsubsection{Unordered set}

Given an (unordered) set of objects $S$, with $|S| = N$, if these are encoded as unsigned integers, we have that the amount of bits required for specifying a single address is $\log N$. The retrieval of any object will bear the same cost. Overall, the total cost of all addresses will be given by:  
$$\mathsf{Total Memory Cost} = N 
\log N$$ 
Therefore, the variety of the observer matches the average memory cost per object:
$$ V =  \log N = \frac{\mathsf{Total Memory Cost}}{N}$$

\subsubsection{Totally ordered set}
Having a linear ordering enables us to perform operations on the addresses. Let us assume that the observer is provided with an ordered set recorded on a stack data structure, whose root address is given. The cost of retrieval corresponds to define how many steps ($0 \le i \le N-1$) we need to go down the stack to find the position of the target object. The total cost of all addresses can be computed as:\footnote{We apply Stirling approximation: $\log(n!) = n \cdot \log n - n + O(\log(n))$.}
$$\mathsf{Total Memory Cost} = \sum_{i=0}^{N-1} \log(i + 1) + \log N = \log N! + \log N \approx N \log N$$
This entails that an ordered data structure is descriptively as efficient as the unordered case (in absolute sense), but opens up to more efficient usage if indexing is made so that elements that are accessed more often have shorter addresses. More importantly, what we observe here is that the variety $V$ of the observer is approximately the same as the average memory cost ($\log N$), just as the unordered case.

\subsection{Entropy vs Variety}

Following Shannon's information theory, entropy captures the average information of an object, seen as being generated from a random source, independently of its internal characteristics. Let us then capture the probability  $p_i$ of object $i$ in terms of Kolmogorov complexity $C_W$ measured on a world machine. To keep the alignment between the two measures of information perfect, we can define this complexity as:
$$ C_W(i) = \log \frac{1}{p_i} $$
We can then rewrite entropy as the sum of generation complexities weighted by probability factors: 
$$H = \sum_{i \in S} p_i \log \frac{1}{p_i} = \sum_{i \in S} p_i C_W(i)$$

In the previous section, by looking at two generic data structures, we have observed that variety can be put in relation to the average cost of retrieval, a notion dual to that of entropy. More generally, rather than cost of retrieval, we can consider the cost of description (which include retrieval operators). We can formally define the cost of describing an object in terms of a Kolmogorov complexity $C_D$ measured on a description machine. We can then redefine variety as the sum of individual descriptive costs of the objects observable by the system, divided per the number of objects:
$$\hat{V} = \sum_{i \in S} \frac{1}{|S|} C_D(i)$$

\subsubsection{Divergence between entropy and variety\label{sec:divergence}
}
Following the standard definitions, entropy is maximal when it coincides with variety, ie. $H_{\max} - V = 0$. Given an arbitrary source defined over the same sample space with entropy $H$, the difference:
$$D = H - V$$ 
should provide a measure of how well the descriptive dimension is, in average, aligned with the generative dimension. It has a negative value, which, at best,  can be equal to 0.\footnote{Information theory refers to a related measure named \textit{redundancy}: $R = 1 - \frac{H}{H_{\max}}$, equal to $-\frac{D}{V}$.} Nevertheless, {taking into account $\hat{V}$ rather than $V$}, we observe that the way the averaging is done differs between entropy $H$ and variety $\hat{V}$: entropy takes into account frequencies of occurrences of objects, variety only the number of objects. This observation suggests that, in order to correctly capture the divergence between the two average measures of information (generative and descriptive), we have to decide upon a more proper common ground for the computation.

\subsection{Three additional perspectives on entropy vs variety\label{sec:three-perspectives}}

By exploring the various possibilities, we discover we can compute the divergence between $H$ and $V$ from three different perspectives: (i) relative to the environment (the world machine), or (ii) absolute (or context-free), or (iii) relative to the observer (the description machine).

\subsubsection{World-relative}
The world-relative perspective (i) gives precedence to the world, therefore the descriptive cost of an object in the divergence is weighted depending on usage, ie. its appearance to the observer, rather than the number of symbols:
$$D_{wrel} = \sum_{i \in S} p_i \left[C_W(i) - C_D(i) \right] $$
Supposing we associate the generation and description machine respectively to the distributions $\mathcal{W}$ and $\mathcal{D}$, $D_{wrel}$ can be described in terms of Kullback-Leibler divergence as:  
$$D_{wrel} = H(\mathcal{W}) - H(\mathcal{W}, \mathcal{D}) = - \mathbb{D}_{\text{KL}}(\mathcal{W} \parallel \mathcal{D})$$
Following Information Theory, a Kullback-Leibler divergence is always positive; we confirm therefore that, as was the case for $D$, $D_{wrel}$ is negative, and is at best equal to 0 when $\mathcal{D}$ is equal to $\mathcal{W}$, corresponding to an optimal encoding scenario.

\subsubsection{Absolute}
The absolute perspective (ii) considers as comparison ground a virtual environment with maximal entropy, ie. whose probability follows a uniform distribution $\mathcal{U}$. Note also that the proposed redefinition of variety can be rewritten as a cross-entropy:
$$\hat{V} = H(\mathcal{U}, \mathcal{D})$$
The divergence given by an absolute perspective would then correspond to:
$$D_{abs} = \sum_{i \in S} \frac{1}{N} \left[C_W(i) - C_D(i) \right] = H(\mathcal{U}, \mathcal{W}) - H(\mathcal{U}, \mathcal{D}) = \mathbb{D}_{\text{KL}}(\mathcal{U} \parallel \mathcal{W}) -  \mathbb{D}_{\text{KL}}(\mathcal{U} \parallel \mathcal{D})$$
This value captures how biased is the world (with respect to a uniform distribution), relatively to how biased is the mind. 

\subsubsection{Mind-relative}
Giving priority to the description machine, we can define  weights for the average in the opposite way than what we did with the world machine, ie. starting from descriptive complexity:
$$d_i = 2^{-C_D(i)}$$
Assuming that these weights satisfy the criteria for a probability distribution over the source $\mathcal{D}$, we can provide a weighted measure of variety (in contrast to the previous one, uniform), defined as:
$$\hat{V}^* = \sum_{i \in S} d_i C_D(i) = H(\mathcal{D})$$
This divergence, measured from a mind-relative perspective (iii), can then be expressed as the following Kullback-Leibler divergence:
$$D_{drel} = \sum_{i \in S} d_i \left[C_W(i) - C_D(i) \right] = H(\mathcal{D}, \mathcal{W}) - H(\mathcal{D}) = \mathbb{D}_{\text{KL}}(\mathcal{D} \parallel \mathcal{W})$$
This result clearly does not match our previous observation on $D$ being negative (section \ref{sec:divergence}), because a Kullback-Leibler divergence cannot be negative. The constraint of negativity was however set by reasoning in terms of maximal entropy, and thus placing ourselves in the context in which the world $\mathcal{W}$  was the source distribution. Here,  the source distribution is instead $\mathcal{D}$ (capturing the mental accessibility to objects), which is observed by means of the model distribution $\mathcal{W}$ (respectively, the world accessibility), which provides a different semantics to the measured divergence. It is sufficient to extract a concept which is conceptually accessible but physically impossible (eg. a unicorn) to increase $D_{drel}$ greatly.

\subsection{Divergences as measures of completeness and soundness}

The complementary functioning of $D_{wrel}$ and $D_{drel}$ can be illustrated with two examples. When an object is common in the world, but very complex to be mentally accessed by the observer (eg. a Belgian Malinois dog, rather common but still conceptually complex for non-experts), its contribution would highly decrease $D_{wrel}$ (becoming more negative), with $D_{drel}$ remaining essentially the same. $D_{wrel}$ captures the discrepancy of the world distribution from the mind distribution, ie. to what extent the world ``fits into'' the mind. In analogy with the notion used with formal systems,\footnote{Formal systems usually refer to distinct definitions of ``truth'' ($\models$) and ``provability'' ($\vdash$). A proof system is \textit{complete} if everything that is true has a proof: if $\phi \models \psi$ then $\phi \models \psi$. A proof system is \textit{sound} if everything that is provable is in fact true: if $\phi \models \psi$ then $\phi \models \psi$.} we can relate a value of $D_{wrel}$ close to 0 to the satisfaction of a sort of \textbf{completeness} criterion:
\begin{quote}\textit{If an object can be easily generated (by the world), then it can be easily described (by the mind).}\end{quote} 
{More formally, this relation can be written as:
$$D_{wrel} \sim 0 \quad \Leftrightarrow \quad \textrm{(for each object)} \quad C_W \sim 0 \rightarrow C_D \sim 0 \; \Leftrightarrow \; C_D \gg 0 \rightarrow C_W \gg 0 $$}

Vice-versa, when an object is easily accessible from a mental point of view, but occurring very rarely in the world (eg. a unicorn), it would contribute to a strong increase to $D_{drel}$ (becoming more positive), while $D_{wrel}$ would not change much. Indeed, $D_{drel}$ captures a discrepancy of the descriptive distribution from the world distribution, i.e. to what extent the mind ``fits into'' the world. A mind which is able to easily generate unicorns would have a high $D_{drel}$. A value of $D_{drel}$ proximate to 0 can be related to the satisfaction of a \textbf{soundness} criterion:
\begin{quote}\textit{If an object can be easily described (by the mind), then it can be easily generated (by the world).}\end{quote} 
{The associated relation is dual to the previous one:
$$D_{drel} \sim 0 \quad \Leftrightarrow \quad \textrm{(for each object)} \quad C_D \sim 0 \rightarrow C_W \sim 0 \; \Leftrightarrow \; C_W \gg 0 \rightarrow C_D \gg 0 $$}

Both completeness and soundness are desired properties for inferential systems. They depend on complementary aspects, not directly transferable from one dimension to the other. {According to the proposed formulas, optimal encoding favours completedness, as common objects would be associated to addresses of minimal length; unexpectedness (as ``being simpler to describe than to generate'') emerges instead as the negation of the soundness criterion at instance level.}

\subsection{Unexpectedness, beyond probability theory}

The formulas introduced above to compute entropy, variety, and the various divergences are based on the assumption of having access to {a whole} sample space $S${, defined both in the ``world'' and the ``mind'' domains.} However, extensionality is easy for finite dictionaries of symbols, or measurable spaces, but not for arbitrarily composable spaces.
{Furthermore, the mind machine may use elements which are non-sensical in themselves with a world machine; for instance, colour is a property characterizing physical objects, it cannot exist standalone.
} 
Yet, an aggregated (even if approximated {or simplified}) view may be useful for {designing or analyzing} batch learning algorithms. 

In all three perspectives {presented in section \ref{sec:three-perspectives}},  unexpectedness plays a fundamental role, because all divergences $D_{wrel}$, $D_{abs}$, $D_{drel}$ are weighted averages of $U$. Minimizing $U$ (e.g. by means of learning) would minimize all three; yet, given the different weights, each perspective provides a different priority on where to intervene first through learning. The potential of this theoretical framework for the design or the interpretation of  machine learning algorithms is yet to be investigated. Nevertheless, from a cognitive standpoint, we already know that a descriptive system is always required to determine an object, before being able to evaluate how frequently this object appears. This entails that modifications should occur first at the level of $\mathcal{D}$ and then have consequences on $\mathcal{W}$, which is in line with what we suggested with the second conjecture. 

Interestingly, the ``physical'' (world-relative) divergence captured by $D_{wrel}$ is in aggregate always negative, because in most cases objects will be irrelevant to be described ($U < 0$). This seems to be dual to what discovered in physics with the second principle of thermodynamics (entropy, in aggregate, always grows in the universe). However, just as in physics entropy can be locally negative, dually, this  divergence can be positive in some local condition, which is precisely where we measure positive unexpectedness ($U > 0$).  

\section{Conclusions}
This document provided a short overview on theoretical elaborations on unexpectedness we are currently exploring. The first conjecture finds formal correspondences between Bayes' rule and the formula of unexpectedness. If these are accepted, unexpectedness may be introduced whenever Bayes' rule is used (eg. variational autoencoders based on ST). On a more fundamental  level, the second conjecture elaborates on the relation between ergodicity and the  causal information associated to situations; we suspect further developments of this theory may contribute to research on the extraction of causal relations. The third conjecture extends the classic definition of variety given in cybernetics, making it dual to that of entropy given in Shannon's information theory. With this transformation, unexpectedness appears to capture a positive local divergence between what the environment provides (generates), and what the observer interprets (decribes), {and corresponds to a measure of local unsoundness of the inferential system.} We hypothesize that taking an aggregated view on these mechanisms can assist in the design of more effective learning processes.

Even if preliminary, these stubs offer a glance of potential directions of novel research. For its grounding in Kolmogorov complexities, the framework offered by ST unveils computational alternatives to probability-based approaches for various inferential tasks, not requiring the reference to prior probabilities, to finite/measurable spaces, and in principle enabling forms of hybrid inference (symbolic and sub-symbolic).

\bibliography{library}

\begin{thebibliography}{1}

\bibitem{Ashby1956}
W.~R. Ashby.
\newblock {\em An introduction to cybernetics}.
\newblock Chapman and Hall, 1956.

\bibitem{Dessalles2011}
J.-L. Dessalles.
\newblock {Coincidences and the encounter problem: A formal account}.
\newblock {\em Proceedings of CogSci 2008: Annual Conference of the Cognitive
  Science Society}, pages 2134--2139, 2008.

\bibitem{Dessalles2008}
J.-L. Dessalles.
\newblock {\em {La pertinence et ses origines cognitives}}.
\newblock Hermes-Science, 2008.

\bibitem{Dessalles2011a}
J.-L. Dessalles.
\newblock {Simplicity Effects in the Experience of Near-Miss}.
\newblock {\em Proceedings of CogSci 2011: Annual Conference of the Cognitive
  Science Society}, pages 408--413, 2011.

\bibitem{Dessalles2013}
J.-L. Dessalles.
\newblock {Algorithmic simplicity and relevance}.
\newblock {\em Algorithmic probability and friends}, 7070 LNAI:119--130, 2013.

\bibitem{Li2008}
M.~Li, P.~Vit{\'a}nyi, et~al.
\newblock {\em An introduction to Kolmogorov complexity and its applications}.
\newblock Springer, 2008.

\bibitem{Shannon1948}
C.~E. Shannon.
\newblock A mathematical theory of communication.
\newblock {\em The Bell system technical journal}, 27(3):379--423, 1948.

\bibitem{Sileno2022}
G.~Sileno and J.-L. Dessalles.
\newblock Unexpectedness and {Bayes’} rule.
\newblock In {\em Proceedings of 3rd International Workshop on Cognition:
  Interdisciplinary Foundations, Models and Applications (CIFMA 2021), joint
  with Software Engineering and Formal Methods. SEFM 2021.}, pages 107--116.
  Springer, 2022.

\end{thebibliography}
\bibliographystyle{abbrv}

\end{document}